\documentclass[runningheads]{llncs}
\usepackage{graphicx}
\usepackage{amsfonts}
\usepackage{bbm}
\usepackage[colorlinks=true,linkcolor=blue, citecolor=blue, urlcolor=blue]{hyperref}
\usepackage{hhline}
\usepackage{ulem}
\usepackage{enumitem}
\usepackage{amsmath}
\usepackage{bbm}
\usepackage{booktabs}  
\usepackage{float}
\usepackage{colortbl}
\usepackage[table]{xcolor}
\usepackage{multirow}
\renewcommand{\labelitemi}{$\circ$}

\begin{document}

\title{PanDx: AI-assisted Early Detection of Pancreatic Ductal Adenocarcinoma on Contrast-enhanced CT}
\titlerunning{PanDx: Early PDAC detection on CECT}

\author{
  Han Liu\thanks{Corresponding author: \texttt{han.liu@siemens-healthineers.com}} \and
  Riqiang Gao \and
  Eileen Krieg \and
  Sasa Grbic
}
\authorrunning{H. Liu et al.}

\institute{
  Digital Technology and Innovation, Siemens Healthineers\\
  \raisebox{-0.2em}{\includegraphics[height=1em]{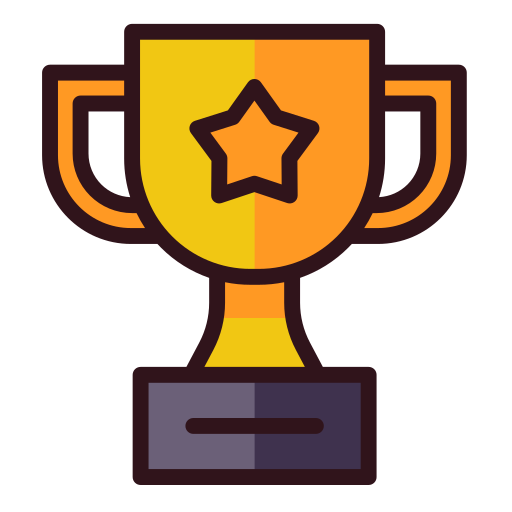}} \textbf{1\textsuperscript{st} place in the PANORAMA Challenge}
}

\maketitle              
\begin{abstract}
Pancreatic ductal adenocarcinoma (PDAC) is one of the most aggressive forms of pancreatic cancer and is often diagnosed at an advanced stage due to subtle early imaging signs. To enable earlier detection and improve clinical decision-making, we propose a coarse-to-fine AI-assisted framework named \textbf{PanDx} for identifying PDAC on contrast-enhanced CT scans. Our approach integrates two novel techniques: (1) distribution-aware stratified ensembling to improve generalization across lesion variations, and (2) peak-scaled lesion candidate extraction to enhance lesion localization precision. PanDx is developed and evaluated as part of the \textbf{PANORAMA challenge}\footnote{Challenge website: \url{https://panorama.grand-challenge.org}}, where it ranked \textbf{1\textsuperscript{st} place} on the official test set with an AUROC of 0.9263 and an AP of 0.7243. Furthermore, we analyzed failure cases with a radiologist to identify the limitations of AI models on this task and discussed potential future directions for model improvement. Our code and models are publicly available at \url{https://github.com/han-liu/PanDx}.

\end{abstract}

\section{Introduction}
Pancreatic cancer (PAC) is the third leading cause of cancer-related mortality, with a dismal 5-year survival rate of approximately 10\% \cite{schwartz2021potential,siegel2019cancer}. The most lethal subtype is pancreatic ductal adenocarcinoma (PDAC), responsible for a disproportionate disease burden, including a 98\% loss of life expectancy and a 30\% increase in disability-adjusted life years \cite{michl2021ueg}. Due to the lack of early and disease-specific symptoms, most PDAC cases are diagnosed at an advanced, non-resectable stage \cite{park2021pancreatic}. If PDAC can be detected at an earlier stage, the quality of life and treatment options for the patients can be substantially improved \cite{singhi2019early,singh2020computerized}. 

Contrast-enhanced computed tomography (CECT) is the most commonly used imaging modality for PDAC detection. Retrospective studies have shown that subtle radiographic signs, such as pancreatic duct cutoff or localized atrophy, can be identified on CECT scans from 3 to 36 months prior to clinical diagnosis \cite{singh2020computerized,toshima2021ct}. However, early-stage PDAC remains difficult to detect due to subtle tumor appearance and absence of secondary signs \cite{elbanna2020imaging}, leading to high inter-reader variability and missed diagnoses. These challenges highlight the need for trustworthy AI systems that can assist radiologists to identify subtle imaging patterns indicative of early PDAC.

PANORAMA is the first challenge for PDAC detection with the largest to-date publicly available dataset, providing a standardized platform to develop and evaluate AI models. In this challenge, participants are provided with a training set of portal-venous CECT scans, along with clinical information and segmentation masks for six PDAC-related structures. The objective is to develop an AI algorithm to predict (1) a PDAC detection map, and (2) a patient-level likelihood score. We have participated in the challenge and developed a two-stage PDAC detection method named \textbf{PanDx}, aiming to segment the target structures in a coarse-to-fine manner. First, we train segmentation models on low-resolution images to efficiently identify the local pancreatic regions. Second, we segment the PDAC-related structures at a finer scale on the cropped regions. Finally, the detection map and the patient-level likelihood are generated based on the segmentation results. Besides, we introduce two novel techniques to further improve the PDAC detection performance, including (1) \textit{distribution-aware stratified ensembling} (DASE) to better generalize across lesion variations, and (2) \textit{peak-scaled lesion candidate extraction} for precise, confidence-guided lesion localization. Evaluated on the official test set of PANORAMA, our method achieved $1\textsuperscript{st}$ place with an AUROC of 0.9263 and an AP of 0.7243.

To better understand model limitations and explore paths toward real-world deployment, we conducted a radiologist-guided failure case analysis. The review revealed common error patterns, including misclassification of cystic lesions and poorly defined tumor margins, suggesting future directions such as incorporating clinical metadata and case-aware learning strategies. In summary, our contributions are as follows:
\renewcommand{\labelitemi}{$\bullet$}
\begin{itemize} 
    \item We present a two-stage PDAC detection framework that achieved \raisebox{-0.2em}{\includegraphics[height=1em]{figures/trophy.png}} \textbf{1\textsuperscript{st} place} on the PANORAMA challenge and release it as an open-source baseline for the research community.
    \item We introduce two complementary techniques: distribution-aware ensembling to improve generalization across lesion types, and peak-scaled lesion extraction for adaptive, confidence-based localization.
    \item We perform a qualitative failure analysis with radiologist input, highlighting common error patterns and providing insights for future model refinement and clinical relevance.
\end{itemize}

\section{Methods}\label{method}

\subsection{Coarse-to-Fine PDAC Detection Framework}\label{overview}

Given the focal and often subtle presentation of PDAC in CECT, we reformulate the detection task as voxel-level segmentation followed by patient-level likelihood estimation. Our PanDx follows a coarse-to-fine design to improve both localization precision and computational efficiency.

\textbf{Stage 1: Pancreas Localization.} We use pretrained 5-fold nnU-Net models \cite{isensee2021nnu} from the PANORAMA baseline \cite{alves2022fully} to segment a combined region consisting of the pancreas, pancreatic duct, and PDAC lesion. These models, trained on low-resolution CECT, are sufficiently robust for localizing the pancreas. Based on these segmentations, we crop a region of interest (ROI) from the original high-resolution scan, using a fixed margin of $100\times50\times15$ mm$^3$.

\textbf{Stage 2: Fine-Scale Structure Segmentation.} Within each ROI, we train another set of nnU-Net models with a ResU-Net backbone to segment all six PDAC-related structures. We replace the default Dice + Cross Entropy (CE) loss with CE loss alone to better align with the detection task \cite{alves2022fully}. The final prediction is obtained by averaging softmax outputs across 5 folds.

\textbf{Detection Score Generation.} We generate a lesion detection map from the softmax output using a modified version of the post-processing algorithm from the Report-Guided-Annotation toolkit \cite{bosma2021annotation}. This method iteratively extracts lesion candidates by identifying the most confident voxel and expanding a connected region based on a confidence threshold. Once the detection map is constructed, we compute the patient-level likelihood score as the maximum voxel-wise probability within the map.

The coarse-to-fine architecture of PanDx offers high-resolution segmentation where needed, while maintaining computational tractability. It also mimics the diagnostic workflow in clinical settings, where radiologists focus detailed attention only on regions of concern.

\begin{figure}[b]
    \centering
    \includegraphics[width=1\linewidth]{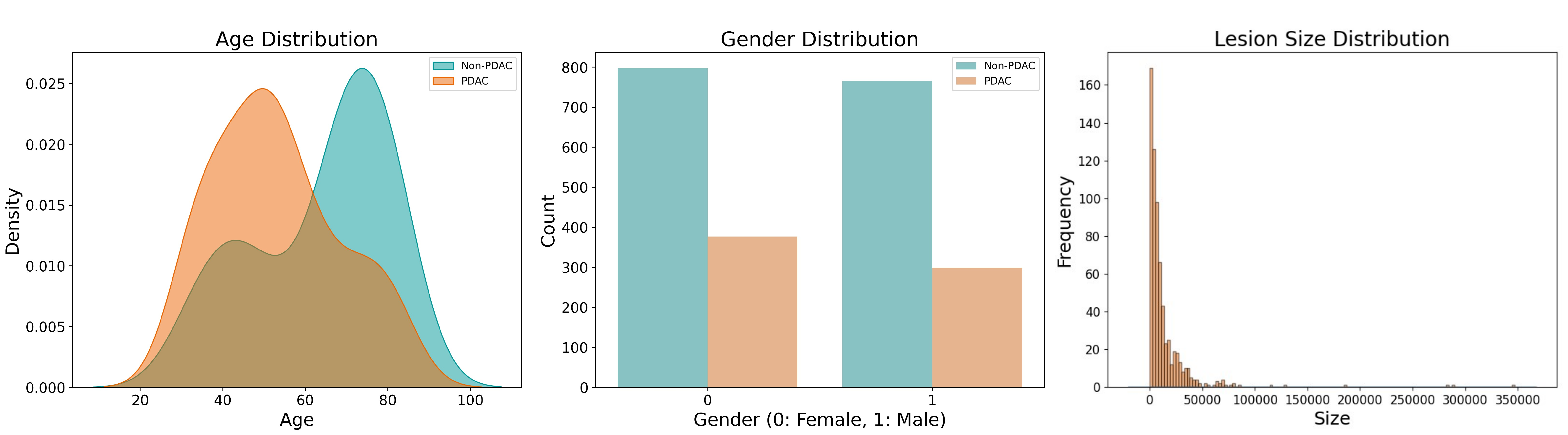}
    \caption{Age (left), gender (middle), and lesion size (right) distributions in the PANORAMA dataset. Lesion size shows the most skew and is used for stratification.}
    \label{fig_dist}
\end{figure}

\subsection{Distribution-Aware Stratified Ensembling (DASE)}\label{novel1}
Robust generalization across diverse patient presentations is critical for real-world deployment of AI systems. In PDAC detection, lesion characteristics can vary significantly in size, shape, and contrast, which poses a challenge for model consistency. To address this, we introduce a stratified ensembling strategy that improves training diversity and promotes performance stability. We analyze three key variables in the challenge dataset: lesion size, age, and gender, as shown in (Fig.~\ref{fig_dist}). Gender shows no clear association with PDAC status, and age exhibits a slight trend that may reflect dataset bias. In contrast, lesion size displays a highly skewed distribution, with a small subset of cases containing very large lesions and the majority containing small ones. 

Because lesion size directly affects detection difficulty and clinical detectability, we use it as the primary factor for stratification. Specifically, we divide PDAC-positive cases into four quartile-based lesion size bins (0–25\%, 25–50\%, 50–75\%, 75–100\%) and distribute them evenly across all five training folds. Additionally, we preserve the global PDAC-to-non-PDAC ratio in each fold to maintain diagnostic class balance. This approach ensures that each model in the ensemble is trained and validated on a representative distribution of lesion types and sizes. By diversifying the training data across folds and stratifying based on clinically meaningful variability, this strategy enhances the ensemble’s ability to generalize across heterogeneous lesions. It also reduces the risk of overfitting to specific lesion subtypes, thereby improving robustness in both internal validation and external deployment.

\subsection{Peak-Scaled Lesion Candidate Extraction}\label{novel2}

Reliable lesion localization is critical for model interpretability and clinical utility. We adapt the region-growing procedure from the Report-Guided-Annotation toolkit \footnote{Toolkit: \url{https://github.com/DIAGNijmegen/Report-Guided-Annotation}} \cite{bosma2021annotation} to extract high-confidence lesion candidates from the softmax probability map \( P(x) \), where \( x \) denotes a voxel in 3D space. The process begins by selecting the most confident voxel, \( x^* = \arg\max_x P(x) \), and growing a candidate region \( C \) by including all connected voxels \( x' \) satisfying \( P(x') \geq \tau \), where \( \tau \) is a confidence threshold. After extracting \( C \), the corresponding voxels are removed from the probability map by setting \( P(x) = 0 \) for all \( x \in C \), and the procedure is repeated to generate multiple lesion proposals.

To account for varying confidence levels across cases, we replace the static threshold with an adaptive strategy defined as \( \tau = \alpha \cdot P(x^*) \), where \( \alpha \) is a scaling factor controlling region expansion relative to peak confidence. We refer to this as \textit{peak-scaled lesion candidate extraction}. This formulation allows the segmentation to adjust contextually to prediction certainty, improving detection reliability and interpretability. We empirically set \( \alpha = 1/15 \) based on cross-validation results (Fig.~\ref{fig_alpha}). Unlike fixed-threshold methods, this scaling factor critically influences the trade-off between sensitivity and specificity. Since optimal \(\alpha\) depends on task characteristics and dataset properties, this adaptive strategy may generalize to other medical applications with heterogeneous lesion morphology and variable prediction confidence.

\section{Experiments and Results}\label{result}
\subsection{Dataset and Evaluation}\label{eval}
The challenge dataset consists of $\sim3000$ portal-venous contrast-enhanced CT scans, where 2238, 86 (30 lesions), and 957 (323 lesions) cases are officially split into training, validation, and testing, respectively. AI algorithms are submitted as Docker containers to the challenge platform for evaluation. Two evaluation metrics are used for final ranking, including (1) Area Under Receiver Operating Characteristic (AUROC) for patient-level diagnosis and (2) Average Precision (AP) for lesion-level detection. The final ranking is computed as the average of the rankings based on AUROC and AP. 

\subsection{Results}\label{result1}
\subsubsection{Challenge Leaderboard}
Tab.~\ref{tab2} summarizes the final testing leaderboard\footnote{\url{https://panorama.grand-challenge.org/evaluation/testing-phase/leaderboard}} on 957 held-out cases. Among all submissions, our PanDx achieved the highest performance in both AUROC (0.9263) and AP (0.7243), ranking $1\textsuperscript{st}$ overall. While several top-performing teams achieved comparable AUROC scores, our method showed a clear advantage in AP, which reflects lesion-level detection quality. The 95\% confidence intervals further support the robustness of our approach, with upper and lower bounds for both AUROC and AP surpassing those of all competing methods. Notably, only three teams outperformed the challenge baseline \cite{alves2022fully}, highlighting the difficulty of the task and the value of our contributions.

% \begin{table}[t]
%     \centering
%     \caption{Testing phase leaderboard (957 cases). \textbf{Bold} indicates the best.}
%     \label{tab2}
%     \rowcolors{5}{white}{gray!15}
%     \begin{tabular}{clcccccc}
%         \toprule
%         \multirow{2}{*}{Rank} & \multirow{2}{*}{Team name} & \multirow{2}{*}{AUROC} 
%             & \multicolumn{2}{c}{95\% CI} & \multirow{2}{*}{AP} & \multicolumn{2}{c}{95\% CI} \\
%         \cmidrule(lr){4-5} \cmidrule(lr){7-8}
%          & & & Lower & Upper & & Lower & Upper \\
%         \midrule
%         1 & Ours
%           & \textbf{0.9263} & \textbf{0.9085} & \textbf{0.9429}
%           & \textbf{0.7243} & \textbf{0.6708} & \textbf{0.7765} \\
%         2 & Hero of Ages
%           & 0.9239 & 0.9068 & 0.9400
%           & 0.6353 & 0.5780 & 0.6910 \\
%         3 & PANORAMiX
%           & 0.9090 & 0.8876 & 0.9288
%           & 0.7004 & 0.6475 & 0.7517 \\
%         4 & FightTumor 
%           & 0.9063 & 0.8849 & 0.9263
%           & 0.6375 & 0.5764 & 0.6965 \\
%         4 & DeepMax
%           & 0.9223 & 0.9046 & 0.9390
%           & 0.6161 & 0.5562 & 0.6736 \\
%         4 & Baseline \cite{alves2022fully}
%           & 0.9223 & 0.9046 & 0.9389
%           & 0.6335 & 0.5756 & 0.6893 \\
%         \bottomrule
%     \end{tabular}
% \end{table}

\begin{table}[t]
    \centering
    \caption{Testing phase leaderboard (957 cases). \textbf{Bold} indicates the best.}
    \label{tab2}
    \rowcolors{5}{white}{gray!15}
    \begin{tabular}{clcccccc}
        \toprule
        \multirow{2}{*}{Rank} & \multirow{2}{*}{Team name} & \multirow{2}{*}{AUROC} 
            & \multicolumn{2}{c}{95\% CI} & \multirow{2}{*}{AP} & \multicolumn{2}{c}{95\% CI} \\
        \cmidrule(lr){4-5} \cmidrule(lr){7-8}
         & & & Lower & Upper & & Lower & Upper \\
        \midrule
        1 & PanDx
          & \textbf{0.9263} & \textbf{0.9085} & \textbf{0.9429}
          & \textbf{0.7243} & \textbf{0.6708} & \textbf{0.7765} \\
        2 & Hero of Ages
          & 0.9239 & 0.9068 & 0.9400
          & 0.6353 & 0.5780 & 0.6910 \\
        3 & PANORAMiX
          & 0.9090 & 0.8876 & 0.9288
          & 0.7004 & 0.6475 & 0.7517 \\
        4 & FightTumor 
          & 0.9063 & 0.8849 & 0.9263
          & 0.6375 & 0.5764 & 0.6965 \\
        4 & DeepMax
          & 0.9223 & 0.9046 & 0.9390
          & 0.6161 & 0.5562 & 0.6736 \\
        4 & Baseline \cite{alves2022fully}
          & 0.9223 & 0.9046 & 0.9389
          & 0.6335 & 0.5756 & 0.6893 \\
        \bottomrule
    \end{tabular}
\end{table}

\subsubsection{Ablation Studies}

To assess the impact of key design choices, we performed ablation studies using the PANORAMA validation leaderboard (86 cases) and internal 5-fold cross-validation. We focus on two components: (1) Distribution-Aware Stratified Ensembling (DASE), and (2) the sensitivity of the scaling factor \(\alpha\) in our peak-scaled lesion extraction.

\textbf{Impact of DASE.}
Our proposed DASE stratifies PDAC-positive cases based on lesion size and maintains this distribution across training folds. We evaluated its impact on the PANORAMA validation set (86 cases). Our experiments show that incorporating DASE improved lesion-level AP from 0.7983 to 0.8247, while AUROC remained comparable (from 0.9798 to 0.9750). This gain reflects increased sensitivity to morphologically diverse tumors. By promoting balanced representation during training and reducing sampling bias, DASE enhances the robustness and generalizability of ensemble models in clinically imbalanced detection settings.

\begin{figure}[t]
    \centering
    \includegraphics[width=1\linewidth]{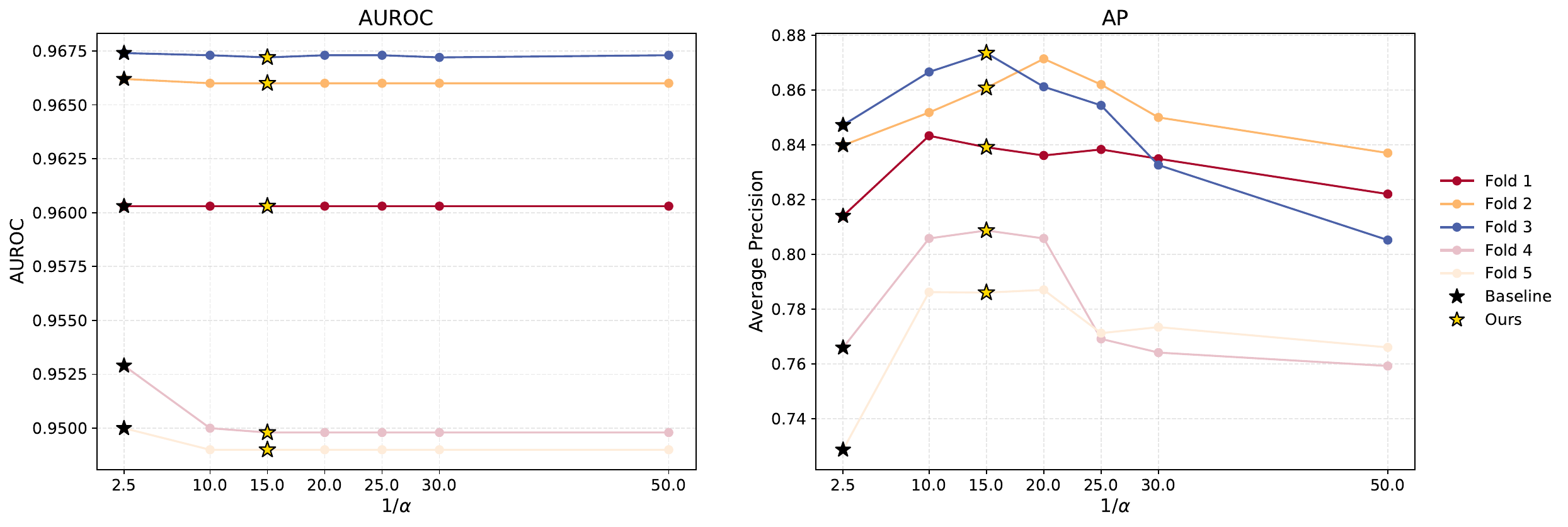}
    \caption{Impact of peak-scaling factors \(\alpha\) on patient-level AUROC and lesion-level AP across 5-fold cross-validation.}
    \label{fig_alpha}
\end{figure}

\textbf{Impact of Peak-scaling factor.}
We investigated the impact of the peak-scaling factor \(\alpha\) with our internal 5-fold cross-validation. As illustrated in Fig.~\ref{fig_alpha}, lesion-level AP reaches its highest performance when \(1/\alpha\) is between 10 and 20, whereas AUROC remains largely stable across different values. When \(\alpha\) is too small, the expansion becomes overly aggressive and includes non-lesion voxels, leading to an increase in false positives. In contrast, smaller \(1/\alpha\) values result in overly conservative region growing, which may undersegment lesions and reduce detection sensitivity. Our empirically selected value of \(1/\alpha = 15\) consistently outperforms the fixed baseline threshold of \(1/\alpha = 2.5\) used in \cite{alves2022fully}. These results emphasize the importance of tuning confidence-adaptive thresholds for effective lesion localization, particularly in applications where lesions are morphologically diverse and often subtle in appearance.

\subsection{Failure Case Analysis}
\begin{figure}[H]
    \centering
    \includegraphics[width=0.95\linewidth]{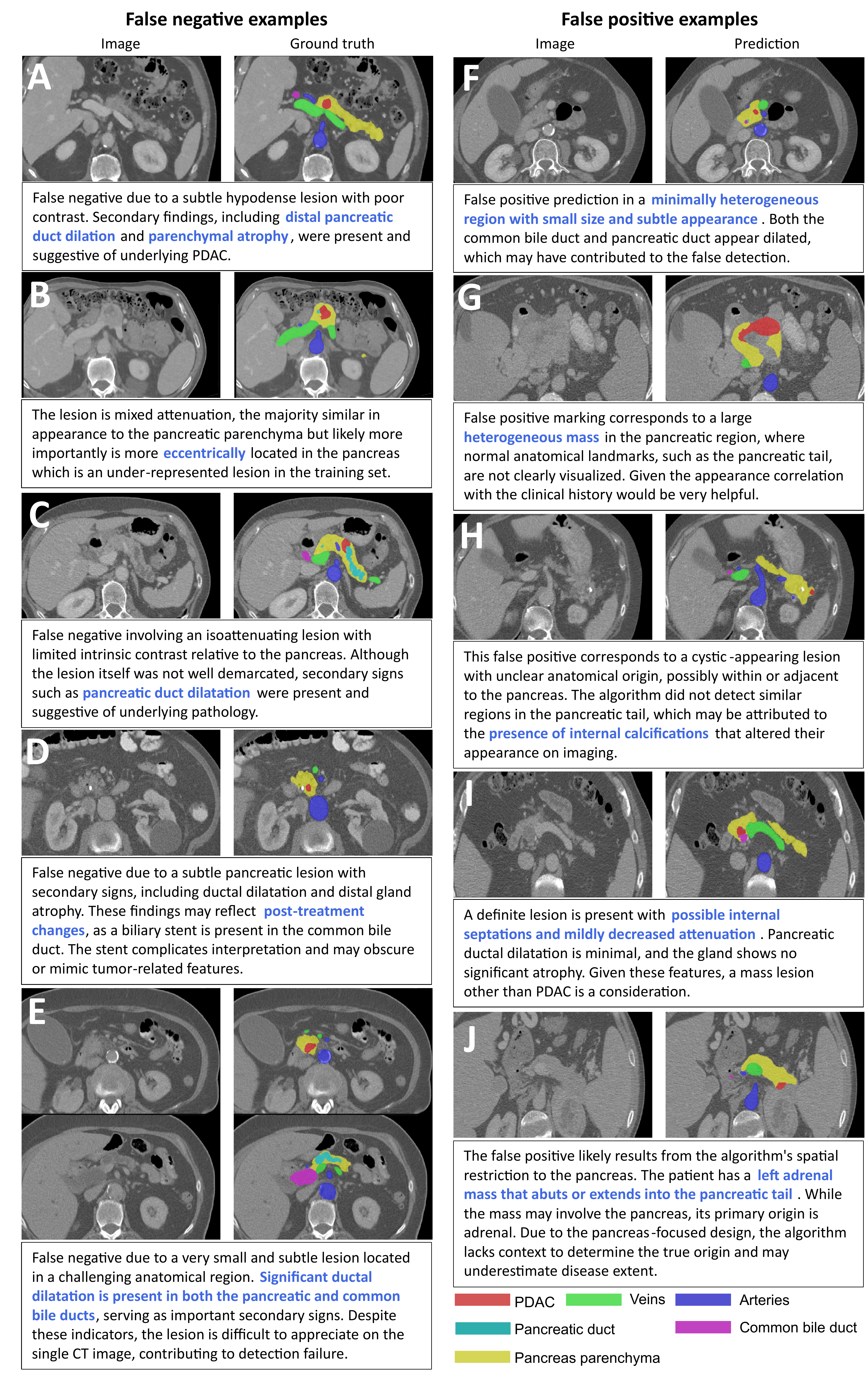}
    \caption{Representative false negative and false positive cases from our model. Each case includes radiologist commentary describing possible factors that may have contributed to model error.}
    \label{fig_fpfn}
\end{figure}

To better understand clinically relevant error modes, we conducted a detailed review of representative failure cases from our final top-ranked solution, in collaboration with a board-certified radiologist. This expert-guided analysis offers valuable insight into the limitations of current deep learning systems for pancreatic cancer detection under real-world clinical conditions. Fig.~\ref{fig_fpfn} presents examples of both false negatives and false positives, each accompanied by radiologist commentary on the underlying causes of model error.

\textbf{False negatives} were commonly associated with lesions that were either subtle in appearance, located in anatomically complex regions, or isoattenuating relative to the surrounding pancreatic parenchyma. As shown in Fig.~\ref{fig_fpfn}A, ~\ref{fig_fpfn}C and ~\ref{fig_fpfn}E, the lesion itself exhibited minimal contrast, making detection challenging for both the model and expert readers. Secondary imaging features, such as pancreatic ductal dilation and distal gland atrophy, were often present and could have served as indirect indicators of malignancy. Moreover, as shown in  Fig.~\ref{fig_fpfn}B, the lesion exhibited mixed attenuation and was eccentrically located within the pancreas, a pattern under-represented in the training set that may have reduced the model’s sensitivity.  Fig.~\ref{fig_fpfn}D showed substantial post-treatment changes, including a biliary stent, which altered the local anatomy and complicated lesion localization.

\textbf{False positives} typically involved areas that appeared mildly heterogeneous or cystic, without clear signs of malignancy. Several cases presented features that visually resembled true lesions but lacked corroborating anatomical changes such as ductal dilation or gland atrophy. As shown in  Fig.~\ref{fig_fpfn}G, the model incorrectly marked a mass arising from the left adrenal gland that was abutting the pancreatic tail. Although anatomically adjacent, the mass was extra-pancreatic in origin, illustrating the model’s limited spatial awareness beyond the pancreas. In  Fig.~\ref{fig_fpfn}I, a lesion with septated cystic characteristics and minimal ductal dilation was flagged, despite the absence of definitive malignant indicators. These findings reflect the difficulty of distinguishing borderline lesions and benign mimics using imaging alone.

This analysis reveals multiple avenues for model improvement. \textbf{First}, explicitly modeling secondary imaging signs such as pancreatic ductal dilation and parenchymal atrophy may enhance sensitivity in cases where the primary lesion lacks clear contrast. \textbf{Second}, expanding the spatial scope of the model to include peripancreatic structures may enhance its ability to discriminate intra- from extra-pancreatic abnormalities. \textbf{Third}, incorporating clinical metadata, such as prior interventions or disease history, may reduce false positives related to treatment effects or post-operative changes. \textbf{Lastly}, diversifying the training data to include atypical or borderline cases could increase model generalizability across a wider clinical spectrum.

\section{Conclusion}
We propose a two-stage framework named PanDx for early PDAC detection in contrast-enhanced CT, achieving $1\textsuperscript{st}$ place on the PANORAMA challenge. Our key contributions include a stratified ensembling strategy and confidence-adaptive lesion extraction. Expert-reviewed failure analysis highlights challenges in AI prediction of isoattenuating tumors and post-treatment anatomy. Future work may incorporate modeling secondary signs, clinical metadata, and tailored training strategies to improve detection of subtle or atypical cases. 

\textbf{Prospect of application} The proposed framework has the potential to support radiologists in the early detection of PDAC by highlighting high-risk regions on routine CECT scans, enabling earlier intervention and improved patient management.

\noindent{\\}\textbf{Disclaimer.} This paper describes research findings not currently available commercially. Future availability is not guaranteed.

\bibliographystyle{splncs04}
\bibliography{references.bib}
\end{document}